\documentclass[letterpaper, 10 pt, conference]{ieeeconf}  

\IEEEoverridecommandlockouts                              

\usepackage{graphicx} 
\usepackage{epsfig} 
\usepackage{times} 
\usepackage{amsmath} 
\usepackage{amssymb}  
\usepackage{cite}
\usepackage{subcaption}
\usepackage{bm}
\usepackage{mathrsfs}
\usepackage{booktabs}
\usepackage{soul}
\usepackage{multirow}
\usepackage[pagebackref,breaklinks,colorlinks,citecolor=blue]{hyperref}
\usepackage{hyperref}
\title{\LARGE \bf
DiffCP: Ultra-Low Bit Collaborative Perception via Diffusion Model
}
\author{Ruiqing Mao$^{1}$, Haotian Wu$^{2}$, Yukuan Jia$^{1}$, Zhaojun Nan$^{1}$, Yuxuan Sun$^{3}$, Sheng Zhou$^{1,*}$,\\ Deniz G{\"u}nd{\"u}z$^{2}$ and Zhisheng Niu$^{1}$
\thanks{$^{*}$Corresponding author: {\tt {sheng.zhou@tsinghua.edu.cn}}%
}
\thanks{$^{1}$Ruiqing Mao, Yukuan Jia, Zhaojun Nan, Sheng Zhou and Zhisheng Niu are with Department of Electronic Engineering, Tsinghua University, Beijing 100084, China.
}
\thanks{$^{2}$Haotian Wu and Deniz G{\"u}nd{\"u}z are with Department of Electrical and Electronic Engineering, Imperial College London, London SW7 2AZ, U.K.
        }
\thanks{$^{3}$Yuxuan Sun is with School of Electronic and Information Engineering, Beijing Jiaotong University, Beijing 100044, China.
       }
}

\begin{document}

\maketitle
\thispagestyle{empty}
\pagestyle{empty}

\begin{abstract}
Collaborative perception (CP) is emerging as a promising solution to the inherent limitations of stand-alone intelligence. However, current wireless communication systems are unable to support feature-level and raw-level collaborative algorithms due to their enormous bandwidth demands. In this paper, we propose DiffCP, a novel CP paradigm that utilizes a specialized diffusion model to efficiently compress the sensing information of collaborators. By incorporating both geometric and semantic conditions into the generative model, DiffCP enables feature-level collaboration with an ultra-low communication cost, advancing the practical implementation of CP systems. This paradigm can be seamlessly integrated into existing CP algorithms to enhance a wide range of downstream tasks. Through extensive experimentation, we investigate the trade-offs between communication, computation, and performance. Numerical results demonstrate that DiffCP can significantly reduce communication costs by 14.5-fold while maintaining the same performance as the state-of-the-art algorithm.
\end{abstract}

\section{Introduction}
The rapid development of intelligent unmanned systems (IUSs) has led to the widespread integration of billions of autonomous devices, such as autonomous vehicles and intelligent robots, into our daily lives. However, single-agent frameworks are constrained by inherent limitations, such as sensor malfunctions, restricted perceptual ranges, and environmental obstructions, which impair their ability to meet the growing demands for safety and reliability, particularly concerning perception range and accuracy.

The evolution of device-to-device (D2D) communication technologies, such as sidelink in cellular vehicle-to-everything (C-V2X) networks, has enabled the sharing of sensing information between agents via wireless channels, a process known as collaborative perception (CP). Prior studies have demonstrated the advantages of CP across various applications~\cite{mao2022dolphins, li2023multi}. However, in non-ideal wireless communication scenarios involving dense deployment, high mobility, and obstructed environments, achieving high reliability and low latency transmission for CP poses significant challenges. Consequently, the maximum achievable data rate in current C-V2X systems is approximately 10 Mbps at a distance of 10 meters, which drops to 5 Mbps at 100 meters~\cite{anwar2019physical}.

\begin{figure}[t!]
    \centering
    \includegraphics[width=1\linewidth]{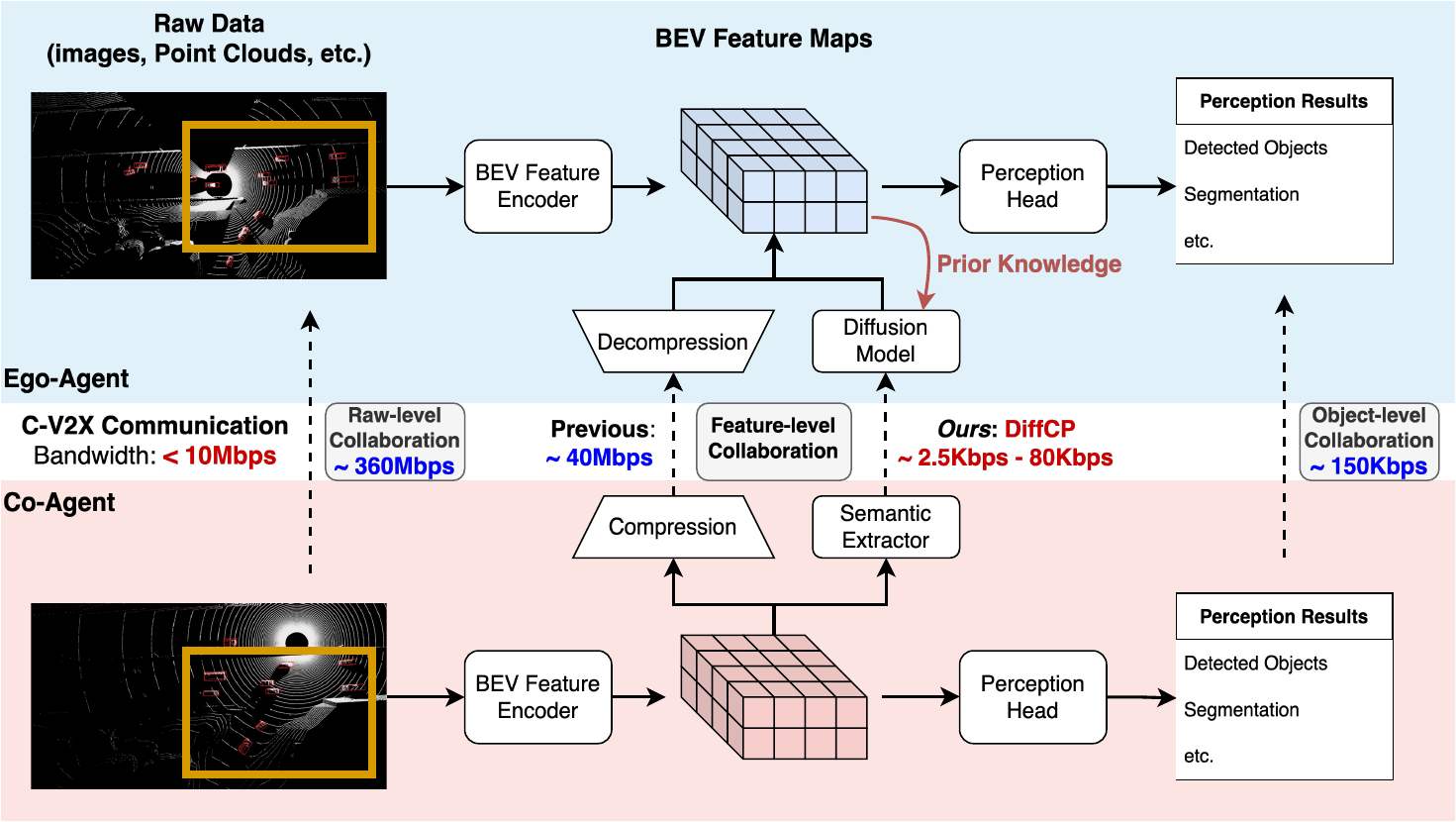}
    \caption{An illustration of CP in IUSs. Compared with previous feature-level CP, the proposed DiffCP leverages the ego-agent's prior geometric to recover feature information from the co-agent. This enables feature-level CP within object-level data rate requirements, satisfying real-world wireless communication constraints.}
    \label{fig:intro}
\vspace{-0.4cm}
\end{figure}

Efforts to reduce the communication cost of perception information have led to the development of various CP architectures\cite{zhou2024task}. As shown in Fig.~\ref{fig:intro}, CP can be broadly classified into three levels based on where the collaborative data is transmitted and fused. Specifically, \textit{raw-level} CP methods~\cite{chen2019cooper} directly exchange raw data to retain detailed information but require substantial bandwidth, demanding approximately 360 Mbps for a 64-line LiDAR and 20 Mbps for a single HD camera, far exceeding C-V2X channel capacities. In contrast, \textit{object-level} CP, like OptiMatch~\cite{song2023cooperative}, reduces bandwidth to around 150 Kbps by transmitting detection results. Nevertheless, it heavily depends on the individual detection capabilities of agents, thereby limiting overall performance. 
Given the importance of intermediate features in perception algorithms, recent research has shifted towards \textit{feature-level} CP. Approaches such as F-Cooper~\cite{chen2019fcooper} and V2VNet~\cite{wang2020v2vnet}, compress raw data into perception-aware features for communication. To address bandwidth constraints, various compression strategies, including channel reduction~\cite{xu2022v2x,xu2022cobevt}, feature downsampling~\cite{coca3d,yang2024how2comm,zhang2024ermvp}, and flow-based~\cite{ffnet} or multi-round selective transmission~\cite{hu2022where2comm}, have been proposed. 
However, these approaches remain limited in either performance or bandwidth efficiency.

In this paper, we propose a novel CP paradigm based on diffusion models, named \textit{\textbf{DiffCP}}, which achieves \textbf{feature-level collaboration} within \textbf{object-level communication costs}. Motivated by the fact that sensor data from collaborative agents can be considered different perspectives of the same scenario, DiffCP leverages the strong correlations between sensor data (as two marked areas in the point clouds in Fig.~\ref{fig:intro}). The key idea behind DiffCP is to answer the question: \textbf{\textit{How many bits are required to characterize the differences between viewpoints}}? Inspired by the success of diffusion models in 3D reconstruction~\cite{liu2023zero}, we employ transformer-based diffusion models~\cite{peebles2023scalable} to capture the geometric and semantic priors of perception information, aiming to reconstruct the observation of the collaborative agent (co-agent) at the ego-agent side. Unlike traditional single-view 3D synthesis, each agent in DiffCP contains unique foreground information due to the occlusions. Therefore, DiffCP incorporates both geometric and semantic conditions into the diffusion model, enabling it to encode the distinctiveness of each viewpoint into a compact feature vector. By performing diffusion within the universal Bird's Eye View (BEV) latent space, DiffCP reduces inference dimensions, supports multiple sensor modalities, and can adapt to diverse downstream tasks. We evaluate the performance of DiffCP on collaborative 3D object detection using the OPV2V dataset. Experimental results demonstrate that DiffCP significantly reduces the communication overhead, even surpassing object-level CP, while maintaining the advantages of feature-level CP, all within a manageable computation time.

In summary, the main contributions of this paper are:

\begin{itemize}
    \item DiffCP is the first CP architecture to use diffusion models to capture both geometric correlations and semantic cues for efficient data transmission. It reconstructs the collaborative sensing information conditioned on the ego-agent's prior knowledge, geometric relationships, and the received semantic features, thereby introducing a new paradigm for CP based on generative models.
    \item DiffCP can be integrated into any existing BEV-based collaborative algorithm for diverse downstream tasks, enabling feature-level collaboration while significantly reducing the bandwidth requirements. This facilitates the implementation and deployment of connected collaborative robotic systems in real-world environments over highly congested wireless networks.
    \item Comprehensive experiments reveal the trade-offs between the reconstruction performance, communication overhead, and computation cost, offering guidance on parameter selection for various IUS scenarios. Numerical results show that DiffCP achieves robust perception performance in ultra-low-bandwidth scenarios, reducing communication costs by 14.5-fold to match the performance of state-of-the-art algorithms.
\end{itemize}

\begin{figure*}[t]
\centering
    \begin{subfigure}[b]{0.48\textwidth}
        \centering
        \includegraphics[width=\textwidth]{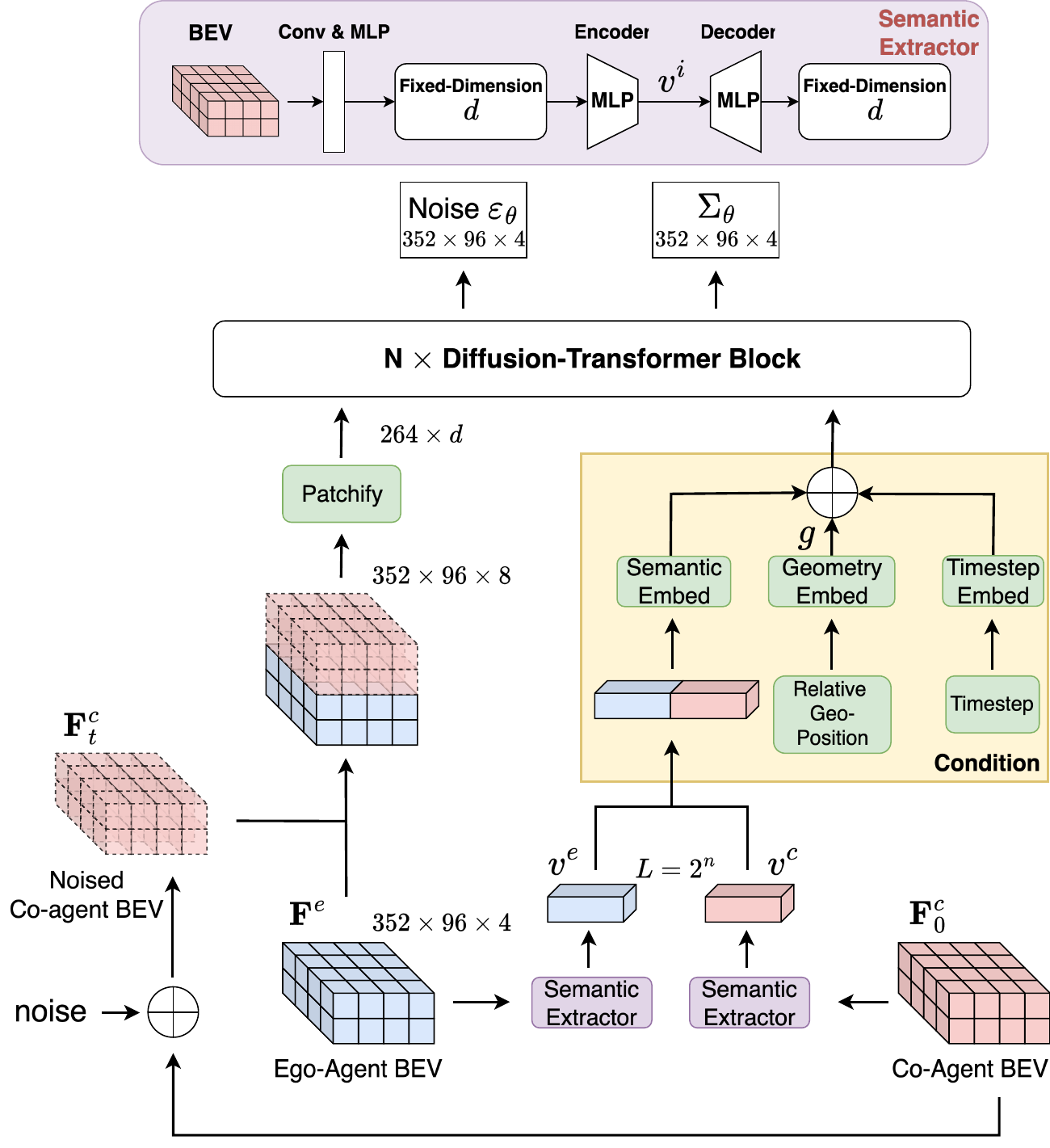}
        \caption{Training process.}
        \label{fig:training}
    \end{subfigure}
    \hfill
    \begin{subfigure}[b]{0.485\textwidth}
        \centering
        \includegraphics[width=\textwidth]{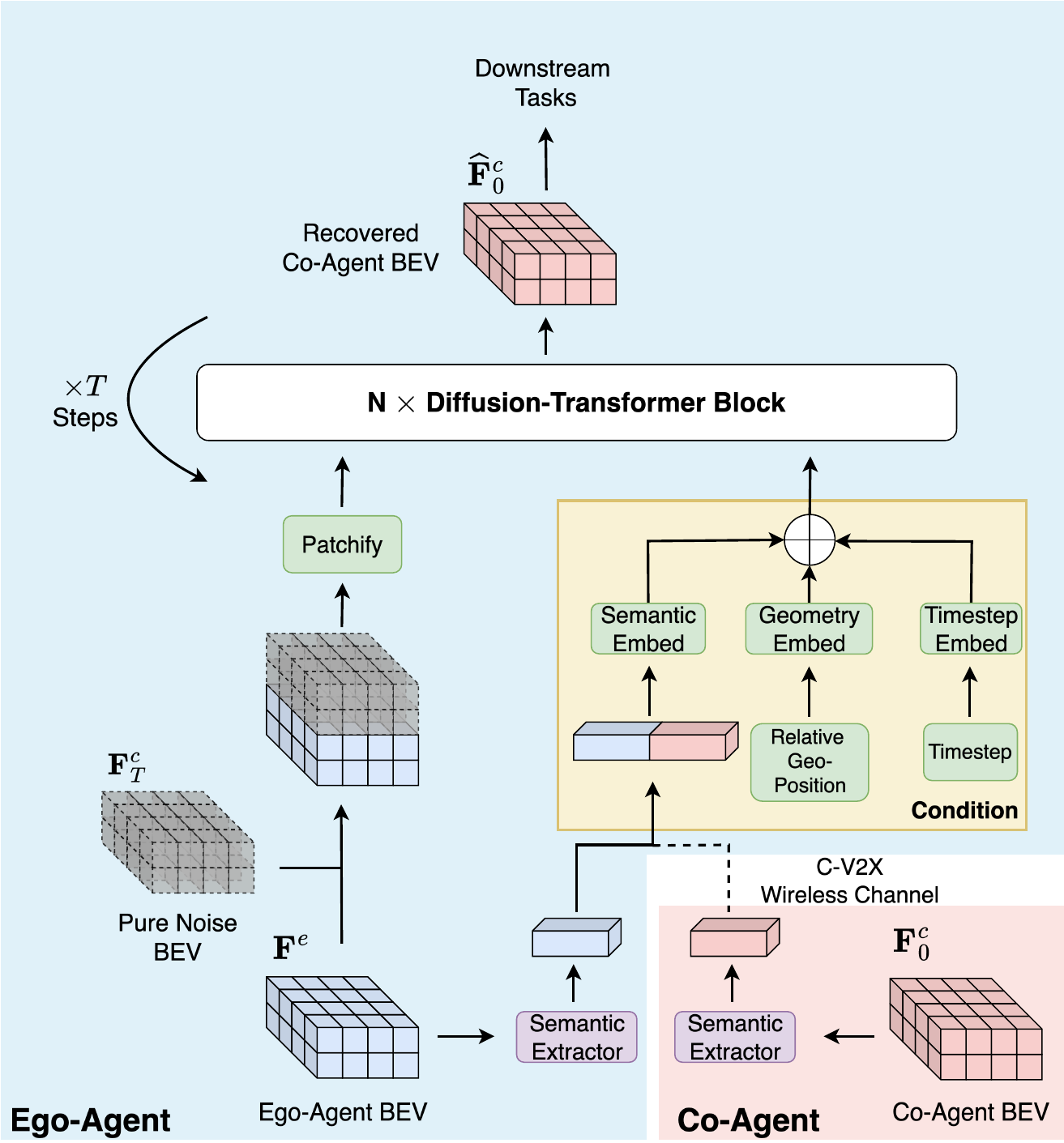}
        \caption{Inference process.}
        \label{fig:inference}
    \end{subfigure}
\caption{The overall architecture of DiffCP. The size of the feature maps is provided as an example to demonstrate the compression functionality. \textbf{Left:}  During the training process, the model is trained using noised BEV features from the co-agent to learn the denoising process. \textbf{Right:}  During the inference process, pure noise features are input to reconstruct the co-agent's BEV features, while only the semantic feature vectors are transmitted through the wireless channel.}
\label{fig:architecture}
\vspace{-0.3cm}
\end{figure*}

\section{Related Work}

\subsection{Collaborative Perception (CP)}
The advantages of connected intelligence have been extensively studied in prior studies across various applications, including object detection~\cite{chen2019cooper}, semantic segmentation \cite{9197364}, and edge inference~\cite{lo2023collaborative}. Several simulated and real-world datasets, such as OPV2V~\cite{xu2022opv2v}, DOLPHINS~\cite{mao2022dolphins}, V2X-Sim~\cite{li2022v2x}, DAIR-V2X~\cite{yu2022dair}, and V2V4Real~\cite{xu2023v2v4real}, have demonstrated the effectiveness of CP in eliminating blind spots, enhancing the perception quality, and reducing the sensing requirements of ego-agents.

However, Ref.~\cite{zhou2024task} highlights the challenges posed by wireless communications over limited bandwidth and time-varying channels, particularly in high mobility scenarios. 
Due to the excessive bandwidth demands of uncompressed feature transmission ($\sim$500 Mbps), various feature compression techniques have been proposed. V2X-ViT~\cite{xu2022v2x} and CoBEVT~\cite{xu2022cobevt} reduce the number of feature channels using convolution layers, while CoCa3D~\cite{coca3d}, MoRFF~\cite{mao2023morff}, How2Comm~\cite{yang2024how2comm} and ERMVP~\cite{zhang2024ermvp} utilize the spatial sparsity of foreground objects to downsample feature maps. FFNet~\cite{ffnet} addresses the temporal correlation of the sensing data, implementing a flow-based transmission scheme. Where2Comm~\cite{hu2022where2comm}, instead, focuses on the spatial correlation between agents and proposes a selective transmission scheme. However, it requires multi-round communications of spatial confidence maps, significantly increasing transmission overhead. Despite outperforming object-level CP and stand-alone intelligence, these methods still require a transmission data rate of around 40 Mbps, exceeding the capacity of real-world wireless communication systems.

\subsection{Diffusion Model}
As an emerging category of generative models, diffusion models~\cite{sohl2015diffusion, ho2020DDPM} have achieved state-of-the-art performance in various vision tasks.
Latent-Diffusion~\cite{rombach2022latent} further reduces computational complexity by operating in latent space. Diffusion-Transformer~\cite{peebles2023scalable} incorporates the transformer architecture to enhance both the scalability and performance of the models. Zero-1-to-3~\cite{liu2023zero} was the first to use diffusion models to capture geometric priors across multi-view images, demonstrating notable success in zero-shot viewpoint transformation from a single input image. However, this approach only addresses single-image 3D view synthesis without incorporating additional information from collaborators.

Recently, diffusion models have been explored in perception tasks~\cite{chen2023diffusiondet,nachkov2023diffusiondetr}. DiffBEV~\cite{zou2024diffbev} exploits diffusion models to enhance the quality of BEV features. DifFUSER~\cite{le2024diffuser} improves robustness in multi-sensor fusion through diffusion processes. BEVWorld~\cite{zhang2024bevworld} employs a latent diffusion model to predict future BEV features conditioned on corresponding actions. However, applications of diffusion models in multi-agent collaborative systems are yet to be investigated.

In collaborative systems, features from different agents often share similar distributions \cite{wu2023features,lin2021completer}, with key differences arising from variations in viewpoint geometries and foreground semantics (i.e., visibility). However, effectively leveraging these correlations under limited communication bandwidth remains an unresolved challenge. Inspired by the powerful conditional generation capabilities of the diffusion models, we employ such a model to characterize the distribution of BEV features conditioned on geometric and semantic correlations. By leveraging this diffusion framework, only the conditioning information needs to be communicated between agents, which enables the ego-agent to reconstruct the co-agent's BEV feature distribution based on its own observations, thereby significantly reducing communication overhead.

\section{Methodology}
We propose DiffCP to reconstruct the co-agent's perception information, i.e., BEV features, at the ego-agent side within a single communication step. The architecture is depicted in Fig.~\ref{fig:architecture}. To compress the feature-level collaborative information, our approach begins by extracting BEV features using pre-trained BEV-based perception algorithms. 
We embed diffusion timesteps, relative geo-positions, and semantic vectors derived from the original BEV features as conditions. These conditions guide both the geometric transformation of ego BEV priors and the infusion of additional information from the co-agent. 
Given that the broadcast of geo-positions has been standardized and widely adopted in the IUSs~\cite{kenney2011dedicated}, the only communication payload needed during feature-level CP deployment is the semantic vector.

\subsection{Data Processing}

As a widely adopted concept, BEV representation provides a global feature space and coordinate system for multi-modality multi-agent collaboration. Our goal is to better reconstruct the co-agent's BEV features at the ego-agent side for all BEV-based CP algorithms, making it a universal CP paradigm.
We illustrate this using CoBEVT~\cite{xu2022cobevt}, one of the current state-of-the-art (SOTA) algorithms for 3D object detection in connected autonomous driving.

Before initiating collaboration, each agent $i$ generates its BEV feature $\textbf{F}^i \in \mathbb{R}^{H\times W\times C}$ from its sensing data, where $H, W, C$ represents the height, width, and the number of feature channels respectively. The dimensions $H$ and $W$ depend on the region of interest (RoI) and the spatial resolution, while $C$ reflects the richness of semantic information. Note that the RoIs generally differ across agents, and the co-agent typically crops its sensing data to match the RoI of ego-agent based on broadcasted geo-positions. In a typical Lidar-based perception setting, each BEV feature tensor is of dimensions $352\times 96\times 256$, resulting in a substantial communication overhead of approximately 33 MB per frame. A simple convolution is adopted to reduce the feature dimension from $256$ to $4$, achieving a 64x compression with slight performance loss. However, the data volume of each BEV feature is still 528 KB, which leads to about 40 Mbps data rate under the typical perception frequency of 10 Hz. In the following sections, we demonstrate how diffusion models can further reduce the data required for transmitting BEV features.

\subsection{Latent BEV Diffusion Formulation}
Inspired by latent diffusion~\cite{rombach2022latent} and diffusion-transformer (DiT)~\cite{peebles2023scalable}, the proposed DiffCP is designed to operate directly in the BEV feature space.
Given the typically large spatial dimensions of pointcloud-based BEV features, we adopt the patchify process from vision transformers (ViTs) with a patch size of $16\times8$, transforming the input BEV feature into a sequence of length $T = \frac{352}{16}\times\frac{96}{8} = 22 \times 12$ with hidden dimension $d$ for the following DiT blocks.
Standard frequency-based positional embeddings in ViT are then applied to these tokens.

Let $\textbf{F}^{e}, \textbf{F}^{c} \in \mathbb{R}^{H\times W\times C}$ denote the BEV feature tensors of the ego-agent and the co-agent, respectively.
Following the typical Gaussian diffusion models, a forward noising process is applied to gradually add noise to the original co-agent BEV features $\textbf{F}^{c}$:
\begin{equation}
q\left(\textbf{F}_t^{c}|\textbf{F}_0^{c}\right)=\mathcal{N}\left(\textbf{F}_t^{c};\sqrt{\Bar{\alpha}_t}\textbf{F}_0^{c},(1-\Bar{\alpha}_t)\textbf{I}\right),
    \end{equation}
where $\Bar{\alpha}_t$ controls the noise strength, and $t$ denotes the timestep. During the training process in Fig.~\ref{fig:training}, the noisy co-agent BEV features can be generated by the reparameterization trick:
\begin{equation}
    \textbf{F}_t^{c} = \sqrt{\Bar{\alpha}_t}\textbf{F}_0^{c} + \sqrt{1-\Bar{\alpha}_t}\epsilon_t, \quad \epsilon_t \sim \mathcal{N}\left(0,\textbf{I}\right).
\end{equation}

The goal of the diffusion model is to learn the reverse denoising process, i.e., the statistics of 
\begin{equation}
p_\theta\left(\textbf{F}_{t-1}^{c}|\textbf{F}_t^{c}\right)=\mathcal{N}\left(\mu_\theta\left(\textbf{F}_t^{c}\right),\Sigma_\theta\left(\textbf{F}_t^{c}\right)\right),
\end{equation} 
where $\mu_\theta$ and $\Sigma_\theta$ denote the original mean and covariance. The loss function is derived through variational lower bound~\cite{kingma2013auto} of the loglikelihood of $\textbf{F}_0^{c}$:
\begin{multline}
 \mathcal{L}(\theta) = -p\left(\textbf{F}_0^{c}|\textbf{F}_1^{c}\right) +\\ 
\Sigma_t \mathcal{D}_{KL}\left(q^*\left(\textbf{F}_{t-1}^{c}|\textbf{F}_t^{c},\textbf{F}_0^{c}\right)\right\| \left. p_\theta\left(\textbf{F}_{t-1}^{c}|\textbf{F}_t^{c}\right)\right).
\end{multline}
Similar to~\cite{peebles2023scalable}, we train the model to predict both the noise $\epsilon_\theta$ and the covariance $\Sigma_\theta$, where the loss function of the noise can be further simplified into:
\begin{equation}
    \mathcal{L}_{simple}(\theta) = \left\| \epsilon_\theta\left(\textbf{F}_t^{c}\right) - \epsilon_t \right\|_2^2.
\end{equation}

During the inference stage (as shown in Fig.~\ref{fig:inference}), the BEV feature is recovered from a pure noise BEV tensor $\textbf{F}_{T}^{c}\sim\mathcal{N}\left(0,\textbf{I}\right)$ through the iterative sampling process $\textbf{F}_{t-1}^{c}\sim p_{\theta}\left(\textbf{F}_{t-1}^{c}|\textbf{F}_{t}^{c}\right)$, where $p_{\theta}$ is the well-trained diffusion model.

\subsection{Learning to Control the Viewpoint}
\label{sec:geo}

To capture the spatial correlation between the ego-agent and the co-agent BEV features, geometric priors must be incorporated into the diffusion process. In real-world collaborative systems, the geometric status of each sensor comprises six elements: $x, y, z$ for its location and $pitch, yaw, roll$ for its orientation. Since the sensing data are projected into the same BEV spatial coordinates, the relative locations are crucial. Specifically, for each pair of collaborative BEV features $\textbf{F}^{c}, \textbf{F}^{e} \in \mathbb{R}^{H\times W\times C}$, the relative geo-position is represented by a vector with three elements: $\bm\Delta=(x^c-x^e, y^c-y^e, z^c-z^e)\in \mathbb{R}^{3}$. A two-layer multilayer perceptron (MLP) network, namely geometry embedder (GE), is then applied to embed the relative geo-position into a vector of length $d$, as follows:
\begin{equation}
        \bm{g} = \text{GE}\left(\bm{\Delta}\right), \bm{g}\in\mathbb{R}^{d}, \bm{\Delta}\in\mathbb{R}^3,
\end{equation}

In order to learn the geometric transformation between agents more effectively~\cite{liu2023zero}, the BEV feature from the ego-agent $\textbf{F}^{e}$ is concatenated with the corresponding noised feature $\textbf{F}_t^{c}$ from the co-agent along the feature dimension before the patchify module, as shown in Fig.~\ref{fig:architecture}. 

\begin{figure*}[htbp]
    \centering
    \begin{subfigure}[b]{0.32\textwidth}
        \centering
        \includegraphics[width=\textwidth]{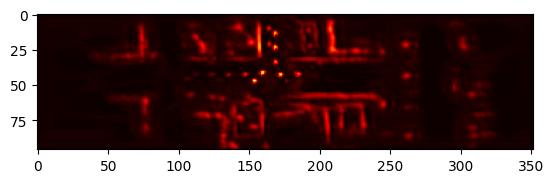}
        \caption{Ego-agent BEV, MSE=0.550}
    \end{subfigure}    
    \begin{subfigure}[b]{0.32\textwidth}
        \centering
        \includegraphics[width=\textwidth]{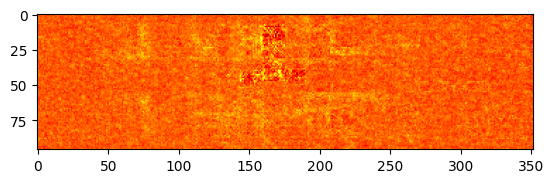}
        \caption{Step=2, MSE=5.507}
    \end{subfigure} 
    \begin{subfigure}[b]{0.32\textwidth}
        \centering
        \includegraphics[width=\textwidth]{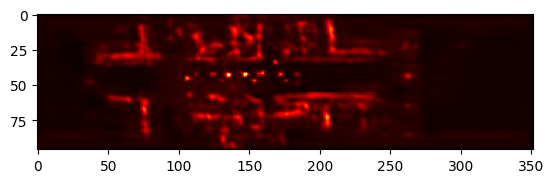}
        \caption{Step=5, MSE=0.271}
    \end{subfigure} 

    \begin{subfigure}[b]{0.32\textwidth}
        \centering
        \includegraphics[width=\textwidth]{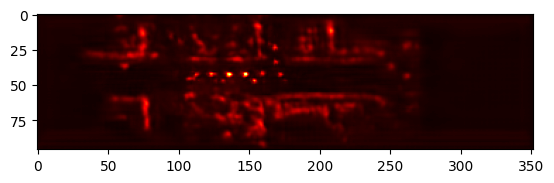}
        \caption{Step=10, MSE=0.285}
    \end{subfigure} 
    \begin{subfigure}[b]{0.32\textwidth}
        \centering
        \includegraphics[width=\textwidth]{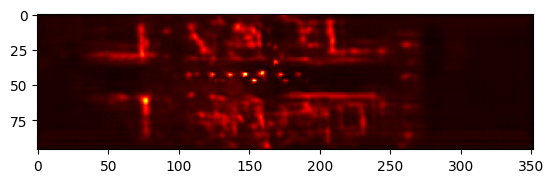}
        \caption{Step=15, MSE=0.294}
    \end{subfigure}
    \begin{subfigure}[b]{0.33\textwidth}
        \centering
        \includegraphics[width=\textwidth]{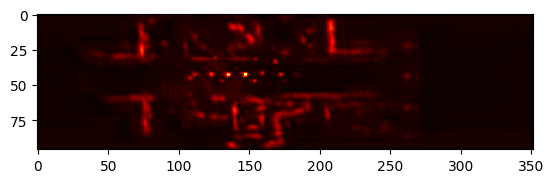}
        \caption{Co-agent BEV, Target}
    \end{subfigure}
\caption{Visualization of the reconstructed BEV features in different sampling steps ($L=512$).}
\label{fig:ddim}
\vspace{-0.4cm}
\end{figure*}

\subsection{Learning to Incorporate Collaborative Information}

Unlike one-to-many view synthesis, co-agents in a collaborative system can provide additional information that helps the ego-agent address blind spots or distant areas.
We introduce a semantic extractor (SE) that generates a unique semantic vector from each BEV feature, which distinguishes between the perceptions of the ego-agent and the co-agent. 
In highly correlated scenarios, minimal additional data from the co-agent are needed to provide information that the ego-agent might not detect or find ambiguous. Since semantic vectors are the sole communication payload in the DiffCP architecture, a reduced vector length means a smaller payload.

To ensure the symmetry between the ego-agent and co-agent, the SEs on both sides share the same parameters. Generally, the SE is composed of several convolution and MLP layers, projecting the input BEV feature to the semantic vector space:
\begin{equation}
    \bm{v}^{\{c,e\}} = \text{SE}\left(\textbf{F}^{\{c,e\}}\right), \bm{v}\in\mathbb{R}^L, \textbf{F}\in\mathbb{R}^{H\times W\times C},
\end{equation}
where the $L$ denotes the vector length. During training, we observed that the model struggles to converge when the vector length is too small. This difficulty may arise because small semantic vectors are insufficient to capture unique information. To address this issue and further reduce the data size, we have redesigned the SE as an encoder-decoder model with a fixed input and output dimension and an adjustable latent dimension. By reducing the vector length, i.e., the latent dimension of SE, most model parameters can be initialized from a previously pretrained model with a larger vector length. This allows us to ``distill" the semantic extraction procedure from a large latent space to a smaller one, enhancing the robustness of model training. 

Together with geometric priors and the typical timestep embedding, these embedded conditions are summed up and fed into the modified adaptive layer norm (adaLN-Zero) module within the DiT block, which calculates scale and shift parameters from the conditions, effectively guiding the denoising process for improved reconstruction quality.

\subsection{Downstream-Aware Augmentation}
\label{sec:augmentation}
As generative models, diffusion models focus on the distribution of the data, rather than the precise values. DiffCP faces a similar challenge. By leveraging geometric priors and additional semantic information, our method significantly reduces communication costs while preserving the feature-level distribution of co-agent perceptions. For tasks where the fine-grained perception is not critical, DiffCP demonstrates competitive performance. However, in tasks that demand exceptionally high precision or strict criteria, recovering the exact values of elements in BEV feature tensors becomes essential.

In this work, we present one solution to adapt DiffCP for a high-precision task, i.e., 3D object detection. Our experiment results reveal that the exact coordinates of 3D bounding boxes for the foreground objects rely on the significant but sparse elements of the BEV feature tensors. By transmitting only the top-K elements, at the expense of extra transmission overhead, the recovered feature maps not only retain the essential distribution of the original perception but also preserve the critical components necessary for accurate bounding box generation. Experiments in Sec.~\ref{sec:experiment} illustrate the effectiveness of this approach as a task-oriented augmentation method.

\section{Experimental Results}
\label{sec:experiment}

\subsection{Experiment Settings}
In this section, we first evaluate the reconstruction accuracy of DiffCP, followed by a discussion on its trade-off between computation time and communication overhead. Next, we apply DiffCP to the 3D object detection task to demonstrate its effectiveness in a specific downstream task.

We adopt the CoBEVT~\cite{xu2022cobevt} as the backbone to extract BEV feature maps from point clouds, which is pretrained on the LiDAR track of OPV2V~\cite{xu2022opv2v}. These BEV feature maps are organized into ego-collaborator pairs, along with their relative positions. We use DiT-L as the backbone of DiffCP ~\cite{peebles2023scalable} with a hidden dimension $d=1024$. The AdamW optimizer is employed with a constant learning rate of $2 \times 10^{-4}$ and a batch size of 512. Each model, with semantic vector lengths $L=512, 256, 128$, is trained on $4 \times$ NVIDIA RTX 3090 graphics cards for 37 hours. Additional training for 13 hours is conducted for fine-tuning models with shorter vector lengths $L=64, 32, 16$.

\begin{table*}[htbp]
\caption{Reconstruction performance and the computation delay under different sampling steps and semantic vector lengths.}
\label{tab:statistic}
\centering
\begin{tabular}{@{}cc|cccccccc@{}}
\toprule
\multicolumn{2}{c|}{Sampling Steps / Computation Time (ms)}         & \multirow{2}{*}{2 / 53} & \multirow{2}{*}{3 / 64} & \multirow{2}{*}{4 / 86} & \multirow{2}{*}{5 / 103} & \multirow{2}{*}{6 / 125} & \multirow{2}{*}{7 / 146} & \multirow{2}{*}{8 / 167} & \multirow{2}{*}{9 / 185} \\ \cmidrule(r){1-2}
\multicolumn{1}{c|}{Semantic Vector Length} & Data Rate &                         &                         &                         &                          &                          &                          &                          &                          \\ \midrule
\multicolumn{1}{c|}{$L=512$}                & 80 Kbps   & 5.507                   & 0.374                   & 0.282                   & \textbf{0.271}           & 0.271                    & 0.275                    & 0.278                    & 0.281                    \\
\multicolumn{1}{c|}{$L=256$}                & 40 Kbps   & 1.820                   & 0.327                   & 0.293                   & \textbf{0.293}           & 0.298                    & 0.305                    & 0.310                    & 0.315                    \\
\multicolumn{1}{c|}{$L=128$}                & 20 Kbps   & 6.095                   & 0.531                   & 0.361                   & 0.339                    & \textbf{0.336}           & 0.339                    & 0.342                    & 0.346                    \\
\multicolumn{1}{c|}{$L=64$}                 & 10 Kbps   & 4.377                   & 0.751                   & 0.439                   & 0.376                    & 0.358                    & 0.352                    & \textbf{0.351}           & 0.351                    \\
\multicolumn{1}{c|}{$L=32$}                 & 5 Kbps    & 3.334                   & 0.588                   & 0.392                   & 0.362                    & 0.355                    & \textbf{0.355}           & 0.358                    & 0.361                    \\
\multicolumn{1}{c|}{$L=16$}                 & 2.5 Kbps  & 6.727                   & 1.572                   & 0.649                   & 0.466                    & 0.404                    & 0.377                    & 0.364                    & \textbf{0.356}           \\ \bottomrule
\end{tabular}
\end{table*}

\subsection{Trade-off Between Computation Time and Reconstruction Accuracy}
In the inference stage, target BEV features are generated through an iterative sampling process. To accelerate the sampling process, we employ DDIM~\cite{song2021denoising} and evaluate feature reconstruction performance using the mean squared error (MSE) between the ground-truth BEV features of the co-agent and the generated ones at the ego-agent. Fig.~\ref{fig:ddim} visualizes the denoising procedure of DiffCP. Starting from a pure noise tensor concatenated with the ego-agent BEV, the model gradually integrates the received semantic information from the co-agent into the ego-agent's observation. This process refines the ego-agent BEV features by removing redundant information and reducing the discrepancy between the generated and target features.  
Table~\ref{tab:statistic} presents the trade-off between the computation time and reconstruction performance. Using the DDIM approach, five sampling steps yield the best quality for models with longer semantic vectors, while shorter vectors require more steps. The entire reconstruction process takes approximately 100 ms, which could be further reduced through software and hardware optimization in real-world deployments.

Interestingly, while more sampling steps should generally improve the performance, excessive sampling steps may increase the influence of ego features. As our goal is to recover the co-agent's BEV features more accurately, redundant information from the ego-agent’s BEV can negatively affect MSE scores. Therefore, a balanced number of sampling steps is necessary to avoid this negative effect. However, in practical collaborative perception, the BEV features will be finally fused into the downstream tasks. The features generated by DiffCP through longer sampling steps might naturally facilitate the fusion of BEV features, effectively integrating the reconstruction and fusion processes into a single module. Future work will explore this integration further.

\subsection{Trade-off Between Data Rate and Reconstruction Accuracy}

As the semantic feature vector is the only communication payload in DiffCP, it is important to clarify the impact of the vector length, i.e., the data rate, on the reconstruction performance. As shown in Table~\ref{tab:statistic}, shorter vector lengths lead to higher MSE scores due to their reduced ability to convey sufficient information, causing instability in the training process. However, our SE architecture, combined with the fine-tuning process, effectively encodes additional information into shorter vectors, resulting in only a 30\% accuracy loss with a 32-fold reduction in length (comparing $L=512$ to $L=16$). Moreover, increasing the number of sampling steps helps maintain promising accuracy at lower data rates, highlighting the compromise between computation time and communication cost.

\begin{table}[t]
\centering
\caption{Comparison of perception performance and data rate requirements of different collaborative 3D object detection algorithms on OPV2V dataset. Data rates are calculated at a typical perception frequency of 10Hz.}
\label{tab:detection}
\begin{tabular}{@{}cccc@{}}
\toprule
Method/Metric            & Data Rate & AP@IoU=0.5 & AP@IoU=0.7 \\ \midrule
No Collaboration                & 0         & 73.25      & 58.22      \\
Object-level              & 219 Kbps  & 86.46      & 79.21      \\
F-Cooper~\cite{chen2019fcooper}                 & 745 Mbps & 87.39      & 79.39      \\
V2VNet~\cite{wang2020v2vnet}                   & 745 Mbps & 91.35      & 82.43      \\
V2X-ViT~\cite{xu2022v2x}                  & 745 Mbps & 91.74      & 83.31      \\
Where2Comm~\cite{hu2022where2comm}               & 745 Mbps & 90.46      & 84.22      \\
ERMVP~\cite{zhang2024ermvp}                    & 745 Mbps & 92.18      & 85.59      \\ \midrule
CoBEVT (compressed)~\cite{xu2022cobevt}                   & 41 Mbps & 90.47      & 84.76      \\
\textbf{+ DiffCP}        & 80 Kbps   & 86.95      & 74.50      \\
\textbf{+ DiffCP + Top-25} & 87.8 Kbps & 89.18      & 78.75      \\ 
+ Top-25 only         & 7.8 Kbps  & 80.73      & 62.14      \\ \bottomrule
\end{tabular}
\end{table}

\subsection{Case Study: 3D Object Detection}
\begin{figure}[]
    \centering
    \includegraphics[width=\linewidth]{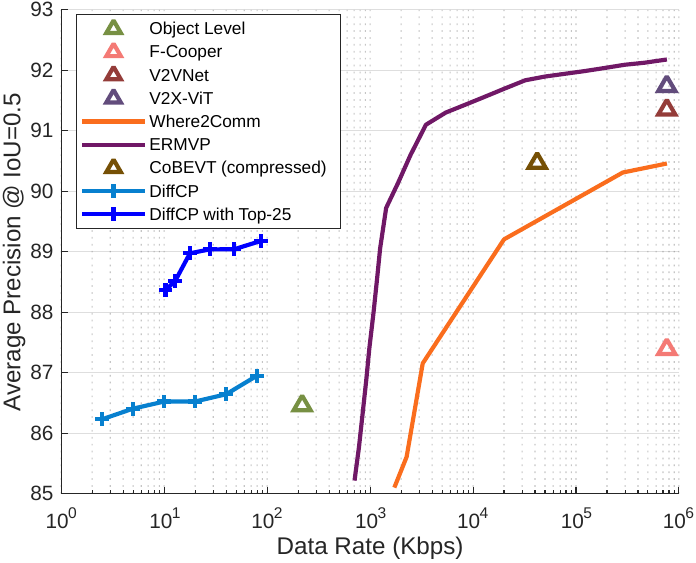}
    \caption{Performance under various data rates.}
    \label{fig:compare}
    \vspace{-0.4cm}
\end{figure}

To further demonstrate the advantages of DiffCP, we evaluate it in collaborative 3D object detection as a case study. We compare DiffCP with several representative feature-level collaborative perception algorithms, including F-Cooper~\cite{chen2019fcooper}, V2VNet~\cite{wang2020v2vnet}, V2X-ViT~\cite{xu2022v2x}, Where2Comm~\cite{hu2022where2comm}, ERMVP~\cite{zhang2024ermvp}, and the original CoBEVT~\cite{xu2022cobevt}. Table~\ref{tab:detection} and Fig.~\ref{fig:compare} illustrate the data rate requirements and the detection precision under two different criteria. 

Compared with current SOTA algorithms, our approach focuses on achieving more collaboration gains with less communication costs, rather than solely improving the precision. As shown in Fig~\ref{fig:compare}, DiffCP achieves comparable performance to object-level CP while requiring less than 5\% of the data rate (reduced from 219 Kbps to 10 Kbps). However, Table. \ref{tab:detection} demonstrates DiffCP performs better under less stringent criteria, such as IoU at 0.5, compared to the stricter benchmarks. This is because DiffCP emphasizes recovering feature distributions rather than exact object elements.

Notably, DiffCP (with top-25 strategy as described in Sec.~\ref{sec:augmentation}) demonstrate the advantages of feature-level collaboration while maintaining an ultra-low data rate. Specifically, it achieves the same level of precision as the SOTA ERMVP with a $14.5$-fold reduction in data rate (from 1.25 Mbps to 87.8 Kbps, as shown in Fig. \ref{fig:compare}) and up to $8,700$ times compression (from 745 Mbps to 87.8 Kbps, as detailed in Table. \ref{tab:detection}), with only a minimal precision loss. Unlike most existing CP algorithms that operate with fixed data rates and thus have limited adaptability to varying wireless conditions, DiffCP addresses this limitation by enabling adaptive rates through variable semantic vector lengths. Additionally, while other algorithms exhibit significant performance degradation at low data rates, DiffCP excels in ultra-low data rate conditions, making it an ideal solution for challenging scenarios. Ablation study reveals that without recovering the complete BEV feature distribution through DiffCP, transmitting only the top-K elements in BEV features results in a significant performance drop. In summary, our approach significantly reduces communication overhead while providing a flexible data rate mechanism, enhancing adaptability in demanding conditions.

\section{Conclusion}
This paper introduces DiffCP, a novel paradigm for CP that utilizes generative models to compress collaborative information between agents. Our approach employs a conditional diffusion model to facilitate the transmission of co-agent perceptual data as a reconstruction task for the ego-agent. This process involves viewpoint transformation based on geometric conditions, as well as the integration of additional information from semantic conditions. Experimental results demonstrate that DiffCP can reconstruct co-agent BEV features at the ego-agent side by exchanging only semantic vectors at a low cost. As a flexible plugin module for CP algorithms, DiffCP is further evaluated on a 3D object detection task, illustrating its ability to enable feature-level collaboration with ultra-low communication overhead. We believe DiffCP can promote the deployment of connected IUSs over existing wireless communication systems.

\addtolength{\textheight}{-1cm}   



\bibliographystyle{IEEEtran}
\bibliography{IEEEabrv,egbib}

\begin{thebibliography}{10}
\providecommand{\url}[1]{#1}
\csname url@rmstyle\endcsname
\providecommand{\newblock}{\relax}
\providecommand{\bibinfo}[2]{#2}
\providecommand\BIBentrySTDinterwordspacing{\spaceskip=0pt\relax}
\providecommand\BIBentryALTinterwordstretchfactor{4}
\providecommand\BIBentryALTinterwordspacing{\spaceskip=\fontdimen2\font plus
\BIBentryALTinterwordstretchfactor\fontdimen3\font minus \fontdimen4\font\relax}
\providecommand\BIBforeignlanguage[2]{{%
\expandafter\ifx\csname l@#1\endcsname\relax
\typeout{** WARNING: IEEEtran.bst: No hyphenation pattern has been}%
\typeout{** loaded for the language `#1'. Using the pattern for}%
\typeout{** the default language instead.}%
\else
\language=\csname l@#1\endcsname
\fi
#2}}

\bibitem{mao2022dolphins}
R.~Mao, J.~Guo, Y.~Jia, Y.~Sun, S.~Zhou, and Z.~Niu, ``Dolphins: Dataset for collaborative perception enabled harmonious and interconnected self-driving,'' in \emph{Proceedings of the Asian Conference on Computer Vision}, 2022, pp. 4361--4377.

\bibitem{li2023multi}
Y.~Li, J.~Zhang, D.~Ma, Y.~Wang, and C.~Feng, ``Multi-robot scene completion: Towards task-agnostic collaborative perception,'' in \emph{Conference on Robot Learning}.\hskip 1em plus 0.5em minus 0.4em\relax PMLR, 2023, pp. 2062--2072.

\bibitem{anwar2019physical}
W.~Anwar, N.~Franchi, and G.~Fettweis, ``Physical layer evaluation of v2x communications technologies: \textsc{5G NR-V2X, LTE-V2X, IEEE} 802.11 bd, and \textsc{IEEE} 802.11 p,'' in \emph{2019 IEEE 90th Vehicular Technology Conference (VTC2019-Fall)}.\hskip 1em plus 0.5em minus 0.4em\relax IEEE, 2019, pp. 1--7.

\bibitem{zhou2024task}
S.~Zhou, Y.~Jia, R.~Mao, Z.~Nan, Y.~Sun, and Z.~Niu, ``Task-oriented wireless communications for collaborative perception in intelligent unmanned systems,'' \emph{IEEE Network}, 2024.

\bibitem{chen2019cooper}
Q.~Chen, S.~Tang, Q.~Yang, and S.~Fu, ``Cooper: Cooperative perception for connected autonomous vehicles based on 3d point clouds,'' in \emph{2019 IEEE 39th International Conference on Distributed Computing Systems (ICDCS)}.\hskip 1em plus 0.5em minus 0.4em\relax IEEE, 2019, pp. 514--524.

\bibitem{song2023cooperative}
Z.~Song, F.~Wen, H.~Zhang, and J.~Li, ``A cooperative perception system robust to localization errors,'' in \emph{2023 IEEE Intelligent Vehicles Symposium (IV)}.\hskip 1em plus 0.5em minus 0.4em\relax IEEE, 2023, pp. 1--6.

\bibitem{chen2019fcooper}
Q.~Chen, X.~Ma, S.~Tang, J.~Guo, Q.~Yang, and S.~Fu, ``F-cooper: Feature based cooperative perception for autonomous vehicle edge computing system using 3d point clouds,'' in \emph{Proceedings of the 4th ACM/IEEE Symposium on Edge Computing}, 2019, pp. 88--100.

\bibitem{wang2020v2vnet}
T.-H. Wang, S.~Manivasagam, M.~Liang, B.~Yang, W.~Zeng, and R.~Urtasun, ``V2vnet: Vehicle-to-vehicle communication for joint perception and prediction,'' in \emph{Computer Vision--ECCV 2020: 16th European Conference, Glasgow, UK, August 23--28, 2020, Proceedings, Part II 16}.\hskip 1em plus 0.5em minus 0.4em\relax Springer, 2020, pp. 605--621.

\bibitem{xu2022v2x}
R.~Xu, H.~Xiang, Z.~Tu, X.~Xia, M.-H. Yang, and J.~Ma, ``V2x-vit: Vehicle-to-everything cooperative perception with vision transformer,'' in \emph{European conference on computer vision}.\hskip 1em plus 0.5em minus 0.4em\relax Springer, 2022, pp. 107--124.

\bibitem{xu2022cobevt}
R.~Xu, Z.~Tu, H.~Xiang, W.~Shao, B.~Zhou, and J.~Ma, ``Cobevt: Cooperative bird's eye view semantic segmentation with sparse transformers,'' \emph{arXiv preprint arXiv:2207.02202}, 2022.

\bibitem{coca3d}
Y.~Hu, Y.~Lu, R.~Xu, W.~Xie, S.~Chen, and Y.~Wang, ``Collaboration helps camera overtake lidar in 3d detection,'' in \emph{Proceedings of the IEEE/CVF Conference on Computer Vision and Pattern Recognition}, 2023, pp. 9243--9252.

\bibitem{yang2024how2comm}
D.~Yang, K.~Yang, Y.~Wang, J.~Liu, Z.~Xu, R.~Yin, P.~Zhai, and L.~Zhang, ``How2comm: Communication-efficient and collaboration-pragmatic multi-agent perception,'' \emph{Advances in Neural Information Processing Systems}, vol.~36, 2024.

\bibitem{zhang2024ermvp}
J.~Zhang, K.~Yang, Y.~Wang, H.~Wang, P.~Sun, and L.~Song, ``Ermvp: Communication-efficient and collaboration-robust multi-vehicle perception in challenging environments,'' in \emph{Proceedings of the IEEE/CVF Conference on Computer Vision and Pattern Recognition}, 2024, pp. 12\,575--12\,584.

\bibitem{ffnet}
H.~Yu, Y.~Tang, E.~Xie, J.~Mao, P.~Luo, and Z.~Nie, ``Flow-based feature fusion for vehicle-infrastructure cooperative 3d object detection,'' \emph{Advances in Neural Information Processing Systems}, vol.~36, 2024.

\bibitem{hu2022where2comm}
Y.~Hu, S.~Fang, Z.~Lei, Y.~Zhong, and S.~Chen, ``Where2comm: Communication-efficient collaborative perception via spatial confidence maps,'' \emph{Advances in neural information processing systems}, vol.~35, pp. 4874--4886, 2022.

\bibitem{liu2023zero}
R.~Liu, R.~Wu, B.~Van~Hoorick, P.~Tokmakov, S.~Zakharov, and C.~Vondrick, ``Zero-1-to-3: Zero-shot one image to 3d object,'' in \emph{Proceedings of the IEEE/CVF international conference on computer vision}, 2023, pp. 9298--9309.

\bibitem{peebles2023scalable}
W.~Peebles and S.~Xie, ``Scalable diffusion models with transformers,'' in \emph{Proceedings of the IEEE/CVF International Conference on Computer Vision}, 2023, pp. 4195--4205.

\bibitem{9197364}
Y.-C. Liu, J.~Tian, C.-Y. Ma, N.~Glaser, C.-W. Kuo, and Z.~Kira, ``Who2com: Collaborative perception via learnable handshake communication,'' in \emph{2020 IEEE International Conference on Robotics and Automation (ICRA)}, 2020, pp. 6876--6883.

\bibitem{lo2023collaborative}
W.~F. Lo, N.~Mital, H.~Wu, and D.~G{\"u}nd{\"u}z, ``Collaborative semantic communication for edge inference,'' \emph{IEEE Wireless Communications Letters}, vol.~12, no.~7, pp. 1125--1129, 2023.

\bibitem{xu2022opv2v}
R.~Xu, H.~Xiang, X.~Xia, X.~Han, J.~Li, and J.~Ma, ``Opv2v: An open benchmark dataset and fusion pipeline for perception with vehicle-to-vehicle communication,'' in \emph{2022 International Conference on Robotics and Automation (ICRA)}.\hskip 1em plus 0.5em minus 0.4em\relax IEEE, 2022, pp. 2583--2589.

\bibitem{li2022v2x}
Y.~Li, D.~Ma, Z.~An, Z.~Wang, Y.~Zhong, S.~Chen, and C.~Feng, ``V2x-sim: Multi-agent collaborative perception dataset and benchmark for autonomous driving,'' \emph{IEEE Robotics and Automation Letters}, vol.~7, no.~4, pp. 10\,914--10\,921, 2022.

\bibitem{yu2022dair}
H.~Yu, Y.~Luo, M.~Shu, Y.~Huo, Z.~Yang, Y.~Shi, Z.~Guo, H.~Li, X.~Hu, J.~Yuan, \emph{et~al.}, ``Dair-v2x: A large-scale dataset for vehicle-infrastructure cooperative 3d object detection,'' in \emph{Proceedings of the IEEE/CVF Conference on Computer Vision and Pattern Recognition}, 2022, pp. 21\,361--21\,370.

\bibitem{xu2023v2v4real}
R.~Xu, X.~Xia, J.~Li, H.~Li, S.~Zhang, Z.~Tu, Z.~Meng, H.~Xiang, X.~Dong, R.~Song, \emph{et~al.}, ``V2v4real: A real-world large-scale dataset for vehicle-to-vehicle cooperative perception,'' in \emph{Proceedings of the IEEE/CVF Conference on Computer Vision and Pattern Recognition}, 2023, pp. 13\,712--13\,722.

\bibitem{mao2023morff}
R.~Mao, J.~Guo, Y.~Jia, J.~Dong, Y.~Sunt, S.~Zhou, and Z.~Niu, ``Morff: Multi-view object detection for connected autonomous driving under communication and localization limitations,'' in \emph{2023 IEEE 98th Vehicular Technology Conference (VTC2023-Fall)}.\hskip 1em plus 0.5em minus 0.4em\relax IEEE, 2023, pp. 1--7.

\bibitem{sohl2015diffusion}
J.~Sohl-Dickstein, E.~Weiss, N.~Maheswaranathan, and S.~Ganguli, ``Deep unsupervised learning using nonequilibrium thermodynamics,'' in \emph{International conference on machine learning}.\hskip 1em plus 0.5em minus 0.4em\relax PMLR, 2015, pp. 2256--2265.

\bibitem{ho2020DDPM}
J.~Ho, A.~Jain, and P.~Abbeel, ``Denoising diffusion probabilistic models,'' \emph{Advances in neural information processing systems}, vol.~33, pp. 6840--6851, 2020.

\bibitem{rombach2022latent}
R.~Rombach, A.~Blattmann, D.~Lorenz, P.~Esser, and B.~Ommer, ``High-resolution image synthesis with latent diffusion models,'' in \emph{Proceedings of the IEEE/CVF conference on computer vision and pattern recognition}, 2022, pp. 10\,684--10\,695.

\bibitem{chen2023diffusiondet}
S.~Chen, P.~Sun, Y.~Song, and P.~Luo, ``Diffusiondet: Diffusion model for object detection,'' in \emph{Proceedings of the IEEE/CVF international conference on computer vision}, 2023, pp. 19\,830--19\,843.

\bibitem{nachkov2023diffusiondetr}
A.~Nachkov, M.~Danelljan, D.~P. Paudel, and L.~Van~Gool, ``Diffusion-based particle-detr for bev perception,'' \emph{arXiv preprint arXiv:2312.11578}, 2023.

\bibitem{zou2024diffbev}
J.~Zou, K.~Tian, Z.~Zhu, Y.~Ye, and X.~Wang, ``Diffbev: Conditional diffusion model for bird’s eye view perception,'' in \emph{Proceedings of the AAAI Conference on Artificial Intelligence}, vol.~38, no.~7, 2024, pp. 7846--7854.

\bibitem{le2024diffuser}
D.-T. Le, H.~Shi, J.~Cai, and H.~Rezatofighi, ``Diffuser: Diffusion model for robust multi-sensor fusion in 3d object detection and bev segmentation,'' \emph{arXiv preprint arXiv:2404.04629}, 2024.

\bibitem{zhang2024bevworld}
Y.~Zhang, S.~Gong, K.~Xiong, X.~Ye, X.~Tan, F.~Wang, J.~Huang, H.~Wu, and H.~Wang, ``Bevworld: A multimodal world model for autonomous driving via unified bev latent space,'' \emph{arXiv preprint arXiv:2407.05679}, 2024.

\bibitem{wu2023features}
H.~Wu, N.~Mital, K.~Mikolajczyk, and D.~G{\"u}nd{\"u}z, ``Features-over-the-air: Contrastive learning enabled cooperative edge inference,'' \emph{arXiv preprint arXiv:2304.08221}, 2023.

\bibitem{lin2021completer}
Y.~Lin, Y.~Gou, Z.~Liu, B.~Li, J.~Lv, and X.~Peng, ``Completer: Incomplete multi-view clustering via contrastive prediction,'' in \emph{Proceedings of the IEEE/CVF conference on computer vision and pattern recognition}, 2021, pp. 11\,174--11\,183.

\bibitem{kenney2011dedicated}
J.~B. Kenney, ``Dedicated short-range communications (dsrc) standards in the united states,'' \emph{Proceedings of the IEEE}, vol.~99, no.~7, pp. 1162--1182, 2011.

\bibitem{kingma2013auto}
D.~P. Kingma, ``Auto-encoding variational bayes,'' \emph{arXiv preprint arXiv:1312.6114}, 2013.

\bibitem{song2021denoising}
\BIBentryALTinterwordspacing
J.~Song, C.~Meng, and S.~Ermon, ``Denoising diffusion implicit models,'' in \emph{International Conference on Learning Representations}, 2021. [Online]. Available: \url{https://openreview.net/forum?id=St1giarCHLP}
\BIBentrySTDinterwordspacing

\end{thebibliography}

\end{document}